\documentclass[conference]{IEEEtran}
\IEEEoverridecommandlockouts
\usepackage{cite}
\usepackage{amsmath,amssymb,amsfonts}
\usepackage{algorithmic}
\usepackage{graphicx}
\usepackage{gensymb}
\usepackage{textcomp}
\usepackage{amsmath}
\usepackage{hyperref}
\usepackage{xcolor}
\usepackage{multicol}

\usepackage[list=true]{subcaption}

\def\BibTeX{{\rm B\kern-.05em{\sc i\kern-.025em b}\kern-.08em
    T\kern-.1667em\lower.7ex\hbox{E}\kern-.125emX}}
\begin{document}

\title{Generative Multi-Stream Architecture For American Sign Language Recognition
}

\author{
\IEEEauthorblockN{Dom Huh, Sai Gurrapu, Frederick Olson, Huzefa Rangwala, Parth Pathak, Jana Kosecka}
\tt{\{dhuh4, sgurrapu, folson, rangwala, phpathak, kosecka\}@gmu.edu}\\
\normalfont{Department of Computer Science, George Mason University}

}

\maketitle

\begin{abstract}
With advancements in deep model architectures, tasks in computer vision can reach optimal convergence provided proper data preprocessing and model parameter initialization. However, training on datasets with low feature-richness for complex applications limit and detriment optimal convergence below human performance. In past works, researchers have provided external sources of complementary data at the cost of supplementary hardware, which are fed in streams to counteract this limitation and boost performance. We propose a generative multi-stream architecture, eliminating the need for additional hardware with the intent to improve feature richness without risking impracticability. We also introduce the compact spatio-temporal residual block to the standard 3-dimensional convolutional model, C3D. Our rC3D model performs comparatively to the top C3D residual variant architecture, the pseudo-3D model, on the FASL-RGB dataset. Our methods have achieved 95.62\% validation accuracy with a variance of 1.42\% from training, outperforming past models by 0.45\% in validation accuracy and 5.53\% in variance.
\end{abstract}

\section{\textbf{Introduction}}
Many individuals with hearing disabilities communicate using the American Sign Language (ASL). However, this creates a communication barrier between those who understand ASL and those that do not. As background, more than one hundred forms of sign language exist around the world. ASL is prominently used in the United States, Canada, and regions in West Africa and Southeast Asia. In the United States alone ASL ranks as the fourth most used language and there are approximately 500,000 speakers \cite{b13}. Sign languages have complex linguistic rules and rely on various phonemic components and body language.

With major advancements in computer vision and sequence modeling, researchers developed tools to handle applications involving images and time-series data. For static images, \cite{c1} achieved 99.31\% on the Massey University Gesture Dataset 2012. However, performance of sequence modeling algorithms on machine translation for sign language gestures with spatio-temporal motion remains substandard to human baselines. To address this issue, we consider the use of a multi-stream generative convolution network to effectively capture and process information for sign language recognition.

The main contributions of this paper can be summarized as:
\begin{itemize}
  \item We propose a three-stream model that utilizes RGB, Generative Depth, and Motion for machine translation of ASL.
  \item We design and implement a generative adversarial network to dynamically produce depth information in real time.
  \item We introduce a deep 3-dimensional convolutional network model (C3D) residual variant.
\end{itemize}

The rest of the paper is organized as follows.  In Sect. 2 we review the related work on developments in video processing. In Sect. 3 we present our methodology, describe the data, and baseline models. In Sect. 4, we introduce our generative multi-stream architecture models. In Sect. 5, we evaluate the performance of our models compared to baseline models.

\section{\textbf{Related Work}}

Sequential computer vision applications such as action detection and recognition have relied on fundamental deep learning algorithms such as convolution and recurrent feed-forward neural networks discussed in \cite{b1}. In specific, \cite{b2} experiments and adds onto these ideas, stacking the familiar VGG Network with long short-term memory (LSTM) layer, creating the LCRN. \cite{b2} explores various applications that merge ideas in dealing with images and sequential data, defining the standard for handling high dimensional sequential datasets.

In more recent years, \cite{b3} and \cite{b4} demonstrated that utilizing 3-dimensional convolutional operations for video recognition outperforms the convolutional recurrent hybrid models by 7.5\% accuracy on the UCF101 dataset. The model discussed in \cite{b4}, C3D (Convolutional 3D), uses the standard backbone of convolution networks, stacking convolution and pooling layers to feed into fully connected layers. \cite{b4} further analyzed their results and concluded that the C3D architecture and its use of 3D convolution leverage both spatial and temporal features together whereas sequential architectures similar to LCRN leave the spatial features with the convolution and introduce the temporal features as “meta-feature” with the recurrent connection. Thus, the spatial and temporal features are processed in union with the convolution operation.

\cite{b5},\cite{b6} demonstrate significant improvement on the C3D when working with video data by introducing residual connections in novel ways. \cite{b5} specifically introduces a residual connection that splits the spatial and temporal filters, similar to the depthwise separable convolution operation. Their results demonstrate that their pseudo-3D block A outperforms their other implementations and past C3D residual variants.

\cite{b8},\cite{b9} introduce multiple streams of inputs, surpassing the performance of single-stream models. \cite{b9} proposed a three-stream network of C3D models that have the following inputs: RGB, depth, and motion. Inspired by multi-stream networks and in an attempt to avoid the use of atypical sensor hardware, we present a generative multi-stream network that does not require any additional inputs aside from RGB data.

\begin{figure}
  \centering
  \includegraphics[width=0.8\linewidth]{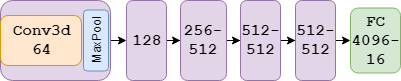}
  \caption{C3D Architecture: The number of filters of each C3D blocks is denoted in the diagram. The initial two C3D blocks have a single convolution layer with 64, 128 filters respectively. The preceding blocks contain multiple convolution layers followed by a pooling layer. Other parameters not specified defined in \cite{b4}.}
  \label{fig:c3d}
\end{figure}

\begin{figure}
  \centering
  \includegraphics[width=\linewidth]{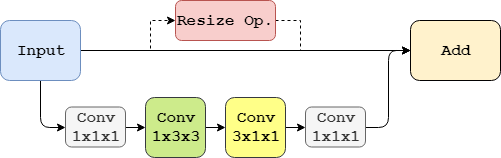}
  \caption{Pseudo-3D Block A: The residual connection consists of four convolution operations. The initial (1x1x1) convolution increases the number of filters to the desired amount. The (1x3x3) convolution acts as a spatial filter, operating solely on the lower dimensions. The (3x1x1) convolution acts as the temporal filter, operating solely at the highest dimension. The final (1x1x1) convolution serves to blend and conclude the residual connection. The resize block linked to the identity connection is not used, but is mentioned for future implementations if resizing occurs; this could be a (1x1x1) convolution for channel upscale or a pooling operation for lower dimension reduction.}
  \label{fig:p3dblock}
\end{figure}

\section{\textbf{Methodology}}
\subsection{\textbf{FASL Dataset}}

\begin{figure}
  \centering
  \includegraphics[width=\linewidth]{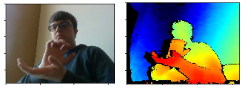}
  \caption{FASL Dataset: A single frame of the word, "movie". The left image is the RGB image, and the right image is the depth image. The dataset can be split into two groups: FASL-RGB, FASL-Depth.}
  \label{fig:fasldata}

\end{figure}

The Frederick American Sign Language (FASL) dataset consists of 1600 videos with ASL signs with the following words: alarm, call, lock, movie, no, off, on, rain, reminder, set, sports, today, tomorrow, weather, yes, and nothing. This dataset was developed by videos captured using Intel RealSense Depth Camera, ranging from 3-10 seconds. This dataset includes depth data mapped to RGB data. The video was segmented into three frames, resulting in 8008 training examples and 5600 validation examples. The 3-frame segmentation was applied after the train-valid split. Various positions and angles were considered while capturing the videos to mimic the behaviors of real ASL users. Lanczos low-pass filter resampling is applied to scale the FASL-RGB image intro appropriate dimensions while preserving the original features. Then the frames were convolved with an edge enhance filter for better feature detection.

Due to the size of the FASL dataset, data augmentation was essential to teach the architecture the desired fidelity and invariance. A random rotation filter was applied to the rotate the images within the range of [-30\degree, 30\degree]. A Gaussian-distributed additive noise filter was also applied to simulate imperfect real-world conditions where the camera lens is covered with dust or has a low resolution. Data augmentation also serves to reduce overfitting and enables the development of a generic ASL recognition model.
\subsection{\textbf{Baseline Models}}
 To validate our multi-stream rC3D architecture, standard machine learning models were implemented to determine the baseline  accuracy for comparison. The dataset was preprocessed by principal component analysis with maximum number of components to maintain over a 98\% variance ratio, in an effort to reduce data dimensionality for these shallow models. Grid search on  hyperparameters was performed on all models to optimize the performance.
 
\begin{table}[htbp]
\caption{Shallow Model Performance}
\begin{center}
\begin{tabular}{|c|c|c|c|}
\hline
\textbf{Shallow}&\multicolumn{3}{|c|}{\textbf{Accuracy(\%)}} \\
\cline{2-4} 
\textbf{Models} & \textbf{\textit{Validation}}& \textbf{\textit{Training}}& \textbf{\textit{Variance}} \\
\hline
Complement Naive Bayes& 26.2 & 54.3 & 28.1\\
\hline
Multinomial Naive Bayes& 26.2 & 48.4 & 22.2\\
\hline
C-SVC& 13.1 & 37.5 & 24.4\\
\hline
Nu-SVC& 53.4 & 71.0 & 17.6\\
\hline
Linear SVC& 53.5 & 71.0 & 17.5\\
\hline
Random Forest& 20.1 & 100  & 79.9\\
\hline
ExtraTree& 26.3 & 100  & 73.7\\
\hline
K-Nearest Neighbors& 46.7 & 51.6 & 4.9\\
\hline
\multicolumn{4}{l}{$^{\mathrm{a}}$On FASL-RGB with PCA}
\end{tabular}
\label{tab1}
\end{center}
\end{table}

 In addition, to validate the new residual connection we introduce to the rC3D model, we developed the P3D-A network using the pseudo-3D block A from \cite{b5}. This network follows the same architecture layout as the C3D.
\begin{figure}
    \centering
    \includegraphics[width=0.8\linewidth]{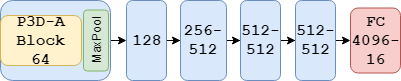}
    \caption{P3D-A Architecture: Similar to the C3D architecture, the 3-dimensional convolution layers are replaced with the Pseudo-3D Block A. Hyperparameters were adjusted to behave like C3D.}
    \label{fig:p3d-a}
\end{figure}

\section{\textbf{Generative Multi-Stream Architecture}}
\begin{figure}
    \centering
    \includegraphics[width=\linewidth]{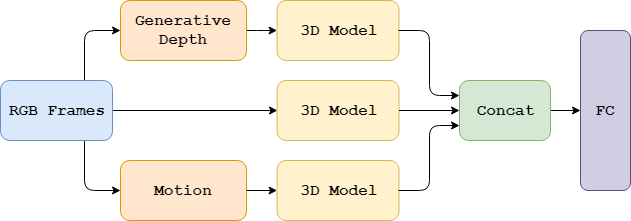}
    \caption{3-Stream rC3D Architecture: The input is the RGB frames, and the other channels are generated and fed into the 3D models. The outputs of the three 3D models are concatenated, pooled, flattened, and fed into the feed-forward model to compute the final output.}
    \label{fig:3st}
\end{figure}

The multi-stream model parallelize the 3D model with different inputs, and each path is called a stream. The architecture can be seen in Figure \ref{fig:3st}. Model parameters are initialized with Kaiming normal. Each affine function is followed with shifted leaky rectified linear unit unless specified. 

The architecture begins by focusing on the high-level details in the first few frames such as the person and the environment and transitions into tracking the salient motion in the subsequent frames such as the movement of the hands.
\subsection{\textbf{3D Model}}

The rC3D  is based on the C3D architecture seen in Fig.\ref{fig:c3d} but replacing the convolutional layers with our residual block, denoted as rC3D blocks. The rC3D residual connection consists of a bottleneck feature, referring to the compression and decompression of the number of filters. The bottleneck was intended to promote generalization in the filters learned.

The spatial pooling was included with the purpose to preserve the temporal dimensions while also reducing the spatial size. With considerations on high computational cost on larger images and maximum feature retention, spatial pooling was performed only in the earlier blocks.

\begin{figure}
\centering
\subcaptionbox{rC3D Block: The pooling operation is only done with the rC3DMP blocks, and it is only performed spatially. The resize block on the identity connection is not used, but is mentioned for future implementations if resizing occurs. Thus a (1x1x1) convolution for channel upscale or a pooling operation for lower dimension reduction may be considered}{\includegraphics[width=\linewidth]{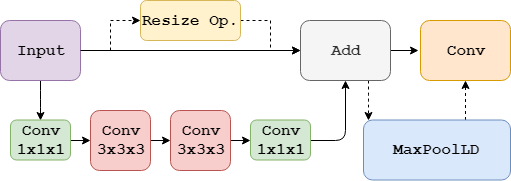}}%
\hfill 
\newline
\subcaptionbox{rC3D Architecture: The number of filters produced by the block is denoted on the rC3D blocks}{\makebox[\linewidth][c]{\includegraphics[width=0.9\linewidth]{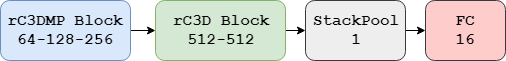}}}
\hfill
\newline
\subcaptionbox{StackPool layer: Concatenates adaptive max and average pooling layers to maximize feature retention with a fixed dimensionality reduction}
{\makebox[\linewidth][c]{\includegraphics[width=.6\linewidth]
    {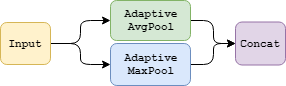}}}
\hfill
\caption{rC3D Network Overview}
\label{fig:rc3d}
\end{figure}

The filters used in the residual connections apply simultaneously on the spatial and temporal dimensions, contrasting the pseudo-3D block, to blend the two aspects of information together rather than incorporating a pipeline. The rC3D architecture depicted in Fig.\ref{fig:rc3d} follows the same practices as C3D architecture in regards to kernel size, padding, and stride.

Low-level motion patterns are learned in the early convolution layers such as moving edges, blobs, edge orientation and color changes. The higher layers learn learn larger motion patterns of body parts, corners, textures, and motion trajectories. The deepest convolutional layers of rC3D learn more complex motion patterns simultaneously with a reference to their spatiotemporal features.

\subsection{\textbf{Generative Depth Model}}

Many cameras do not have a built-in depth sensor. The generative model was intended to dynamically generate the depth information in real time in order to eliminate the need for a camera with a depth sensor. The development and training of the model was supported by FastAI, utilizing their GAN Learner and Switcher.

\begin{figure}
  \centering
\subcaptionbox{Dynamic U-Net Architecture for Depth Generation: This architecture has dense skip connections represented as red lines, otherwise is added if not fed into a block. The number of filters is denoted on the blocks.}{\includegraphics[width=\linewidth]{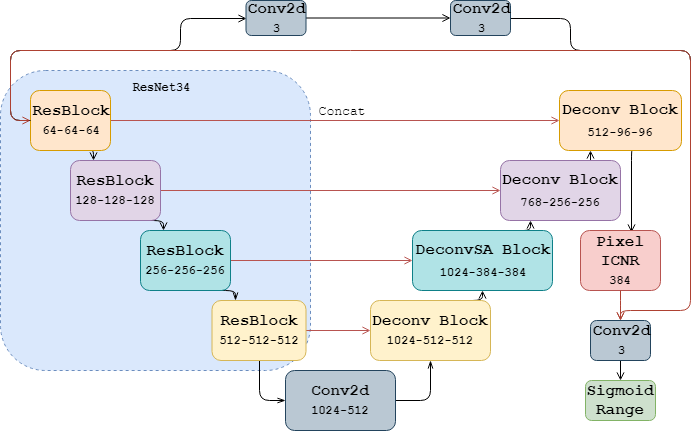}}%
\hfill
\subcaptionbox{Deconv Block: Pixel-ICNR uses the pixel shuffle operation to upscale the downsampled image, then pads the upscaled image with replication padding to finally blur the image with average pooling with kernel size of 2 and stride of 1. Proximal, or nearest, interpolation is applied on the output of the Pixel-ICNR, and is concatenated with its rightful encoder feature map to be convolved. The self-attention convolution is only applied with DeconvSA Block.  }{\includegraphics[width=\linewidth]{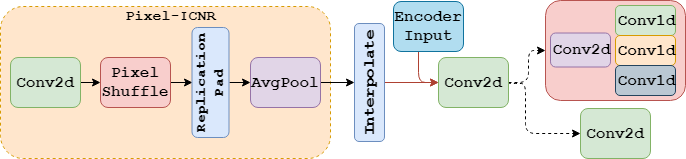}}%
\hfill
\subcaptionbox{Self-Attention Convolutional Architecture as Adversarial Network: DOConv block consists of 2-dimensional dropout passed into 2-dimensional convolutions. The red line connection depicts concatenation. The self-attention convolution is denoted with the key, query, and value convolutions blocked together.}{\includegraphics[width=\linewidth]{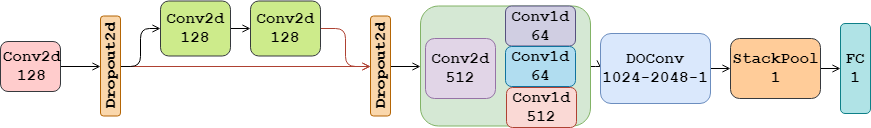}}%
\hfill
\caption{Generative Adversarial Network Overview}
\label{gan}
\end{figure}

The U-Net architecture, derived from \cite{b10}, uses the ImageNet pretrained ResNet34 backbone as the encoder, thus utilizing transfer learning with the generative model for higher performance. The decoder consist of deconvolution blocks, seen in Fig.\ref{gan}. The deconvolutional blocks utilize pixel shuffle with replication padding and proximal interpolation to upscale the image back to the input shape. To truly optimize the generative model, we incorporated the adversarial approach.

Initially, the generative model was trained independently: using Mean Squared Error Loss 

\begin{equation}
MSE = \frac{1}{n}\sum_{i=1}^{n} (y_i  - \hat{y_i})^2
\end{equation}

for each output pixel on the FASL-RGB as the feature \((\hat{y}\textsubscript{i})\) and FASL-Depth as the target \((y\textsubscript{i})\). However, the results lacked many important components of encapsulating the depth information, such as defined segmentation edges and blurred depth colors.

To incorporate the adversarial method, the critic was first trained by optimizing an adaptive binary cross entropy loss on the pre-adversarial generative model’s prediction labeled as fakes and the real depth images labels as reals, achieving 69.7\% validation accuracy on this meta-dataset. The binary cross-entropy loss can be represented as 

\begin{equation}
E(p,q) = -\sum p_i  log(q_i)
\end{equation}
\begin{equation}
E(p,q) = -y  log(\hat{y}) - (1-y)log(1-\hat{y})
\end{equation}

where \(p\) is the set of true labels, \(q\) is the set of prediction, \(y\) is a true label and \(\hat{y}\) is the predicted probability. Finally, the adversarial training consisted training both models in union using FastAI's GAN Switcher. Bilinear interpolation and filtering are applied to the final generative image for better texture and detail reconstruction.

\begin{figure}
\centering
\subcaptionbox{FASL-Depth}{\includegraphics[width=0.3\linewidth]{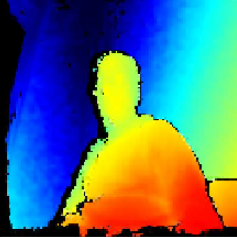}}%
\hfill
\subcaptionbox{Pre-Adversarial}{\includegraphics[width=0.3\linewidth]{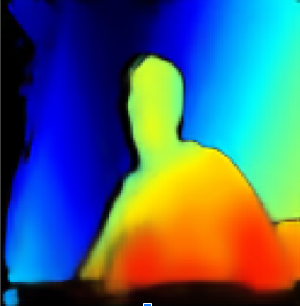}}%
\hfill
\subcaptionbox{Post-Adversarial}{\includegraphics[width=0.3\linewidth]{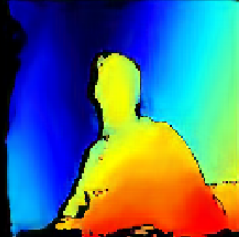}}%
\hfill
\caption{Generative Depth Model: Checkpoint Outputs}
\end{figure}

\subsection{\textbf{Motion}}

 The dense optical flow image illustrates the intensity of the magnitude of the displacement vector fields. The images are generated based on the changes from spatial and temporal dimensions in adjacent frames, utilizing the Gunner Farneback’s algorithm with OpenCV. The motion image is passed as a single frame into the 3D model. 
 
 Farneback's algorithm attempts to compute the optical flow vector for every pixel of each frame \cite{b12}. We can express the image intensity \(I\) as a function of space \((x,y)\) and time \((t)\). Consider a pixel in a frame \(I(x,y,t)\) and it shifts by a distance of \((\delta x,\delta y)\) in the consecutive frame over \(t\) time. Since the pixel intensities are constant between adjacent frames, we can assume
 
\begin{equation}
I(x,y,t) = I(x + \delta x, y + \delta y, t + \delta t).
\end{equation}

 The taylor series approximation of the right-hand side is taken and divided by \(\delta t\) to derive the optical flow equation
 
 \begin{equation}\label{eq:1}
f_{x}u + f_{y}u + f_{t} = 0
\end{equation}

where
\begin{align}
f_{x} = \frac{\delta f}{\delta x} && f_{y} = \frac{\delta f}{\delta y}
\end{align}
\begin{align}
u = \frac{\delta x}{\delta t} && v = \frac{\delta y}{\delta t}.
\end{align}

In the optical flow equation (5) \(f_{x}\), \(f_{y}\), and \(f_{t}\) represent the image gradients along the horizontal axis, vertical axis, and time respectively. The equation is optimized for the two-channel array of optical flow vectors \((u,v)\) and their magnitude and direction are determined to produce the dense optical flow from adjacent frames.

\begin{figure}
  \centering
\subcaptionbox{FASL-RGB}{\includegraphics[width=0.4\linewidth]{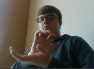}}%
\hfill
\subcaptionbox{Motion Image}{\includegraphics[width=0.4\linewidth]{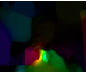}}%
\hfill
\caption{Dense Optical Flow}
  \label{fig:c3d}
\end{figure}

\section{\textbf{Results}}

The developments made on the past C3D with our rC3D model improves validation accuracy by 9\% while reducing variance by 15.84\% when comparing the C3D and the 1-Stream RGB rC3D on FASL-RGB dataset. Even compared to the P3D-A model, the 1-Stream RGB rC3D outperforms the P3D-A model slightly in validation and training accuracy, but improves variance by 5.53\%, confirming the generalization effects of the bottleneck residual connection. However, the compact quality of the residual connection cannot confirmed to improve performance.

The depth and motion features in isolation did not perform well on the rC3D model, which is to be expected as the data is simply complementary to the RGB.
\begin{table}[htbp]
\caption{Model Performances}
\begin{center}
\begin{tabular}{|c|c|c|c|}
\hline
\textbf{Tested}&\multicolumn{3}{|c|}{\textbf{Accuracy(\%)}} \\
\cline{2-4} 
\textbf{Models} & \textbf{\textit{Validation}}& \textbf{\textit{Training}}& \textbf{\textit{Variance}} \\
\hline
C3D*& 76.0  & 99.9  & 23.9  \\
\hline
VGG16-LSTM*& 43.8  & 69.9  & 26.1  \\
\hline
P3D-A*& 84.60 & 98.19 & 13.59 \\
\hline
1-S RGB rC3D* (ours)& 85.05 & 93.11 & 8.06  \\
\hline
1-S Depth rC3D** (ours)& 73.91 & 87.82 & 13.91 \\
\hline
1-S GDepth rC3D* (ours)& 76.35 & 84.32 & 7.97  \\
\hline
1-S Motion  rC3D* (ours) & 74.88 & 83.14 & 8.26  \\
\hline
1-S GPTDepth rC3D* (ours)& 73.36 & 82.42 & 9.06  \\
\hline
2-S RGB-GDepth rC3D* (ours)& 95.39 & \textbf{98.20} & 2.81  \\
\hline
3-S  rC3D* (ours)& \textbf{95.62} & 97.04 & \textbf{1.42} \\
\hline
\multicolumn{4}{l}{$^{\mathrm{*}}$On FASL-RGB, $^{\mathrm{**}}$On FASL-Depth} S = Stream
\end{tabular}
\label{tab1}
\end{center}
\end{table}

The use of the generative model, 1-Stream GDepth rC3D model, compared to the the 1-Stream Depth rC3D model, which learned on FASL-depth, improves validation accuracy by 2.44\% and lowers variance by 5.96\%. This result implies that generative data provide more generalization, and reduce chances of overfitting. To gain insight on the similarity between the generated depth and the actual depth information, the 1-Stream GPTDepth rC3D ran inference on the generated depth frames with the pretrained 1-Stream Depth rC3D. GPTDepth's performance falls within 0.6\% of the performance on the actual depth images, implying high similarity between generative and actual depth data.

The 3-stream models outperformed all models tested on the validation set, but performed worse than the 2-stream RGB-GDepth model on the training set. We infer that this is due to the training conditions we set, which were to limit all the training hyperparameters, including but not limited to learning rate, number of epochs, and optimizer, to be the same, and thus convergence was not met for many of these models. This restriction was set to accurately evaluate our methods against baselines and our results on a dataset with relatively sparse classes.

\section{\textbf{Conclusion}}
In this paper, we try to address the problem of recognizing ASL signs with complex spatiotemporal features on a high-dimensional sequential video dataset. We introduced a multi-stream generative architecture with an improved C3D model for dynamic machine translation of ASL. We demonstrated the usage of multiple streams on feature-rich RGB data and we showed that utilizing generative abilities and motion can increase the performance of ASL sign recognition. In future work we would like to incorporate environment and facial expression recognition for context-based learning to further boost the performance. 

\section{\textbf{Acknowledgements}}
We thank Yuanqi Du and Jay Deorukhkar for helpful comments and discussion.

\bibliographystyle{./bibliography/IEEEtrans}
\bibliography{./bibliography/IEEEabrv,./bibliography/conferencebib}
\end{document}